\documentclass[sigconf]{acmart}
\AtBeginDocument{%
  \providecommand\BibTeX{{%
    \normalfont B\kern-0.5em{\scshape i\kern-0.25em b}\kern-0.8em\TeX}}}

\setcopyright{acmcopyright}
\copyrightyear{2018}
\acmYear{2018}
\acmDOI{XXXXXXX.XXXXXXX}

\acmConference[Conference acronym 'XX]{Make sure to enter the correct
  conference title from your rights confirmation emai}{June 03--05,
  2018}{Woodstock, NY}
\acmPrice{15.00}
\acmISBN{978-1-4503-XXXX-X/18/06}

\usepackage{graphicx}
\usepackage{booktabs}
\usepackage{caption}
\usepackage{multirow}
\usepackage{subcaption}
\usepackage{bbding}
\usepackage{array}
\usepackage{balance}
\usepackage{color}
\usepackage{listings}

\begin{document}

\title{NeuralKG: An Open Source Library for Diverse Representation Learning of Knowledge Graphs}



\author{Wen Zhang$^1$, Xiangnan Chen$^1$, Zhen Yao$^1$, Mingyang Chen$^2$, Yushan Zhu$^2$, Hongtao Yu$^2$, \\Yufeng Huang$^1$, Zezhong Xu$^2$, Yajing Xu$^1$, Ningyu Zhang$^{1,3}$, Zonggang Yuan$^5$, Feiyu Xiong$^6$, \\Huajun Chen$^{2,3,4}$}
\affiliation{%
  \institution{$^1$School of Software Technology, $^2$College of Computer Science and Technology, Zhejiang University \\
  $^3$Alibaba-Zhejiang University Joint Research Institute of Frontier Technologies\\
  $^4$Hangzhou Innovation Center, Zhejiang University \;  $^5$Huawei Technologies Co., Lt \;  $^6$Alibaba Group}
  \streetaddress{}
  \city{}
  \state{}
  \country{}
  \postcode{}
}








\renewcommand{\shortauthors}{Trovato and Tobin, et al.}

\begin{abstract}
NeuralKG is an open-source Python-based library 
for diverse representation learning of knowledge graphs. It implements three different series of Knowledge Graph Embedding (KGE) methods, including conventional KGEs, GNN-based KGEs, and Rule-based KGEs. With a unified framework, NeuralKG successfully reproduces link prediction results of these methods on benchmarks, freeing users from the laborious task of reimplementing them, especially for some methods originally written in non-python programming languages. Besides, NeuralKG is highly configurable and extensible. It provides various decoupled modules that can be mixed and adapted to each other. Thus with NeuralKG, developers and researchers can quickly implement their own designed models and obtain the optimal training methods to achieve the best performance efficiently. We built an website\footnote{Project website: \url{http://neuralkg.zjukg.cn}} to organize an open and shared KG representation learning community. The library, experimental methodologies, and model reimplement results of NeuralKG are all publicly released at \url{https://github.com/zjukg/NeuralKG}.
\end{abstract}



\maketitle

\section{Introduction}
Knowledge Graphs (KGs) represent real-world facts as symbolic triples in the form of \textit{(head entity, relation, tail entity)}, abbreviated as \textit{(h,r,t)}, for example, \textit{(Earth, containedIn, SolarSystem)}. 
Currently, many large-scale knowledge graphs have been proposed, such as YAGO \cite{yago}, Freebase \cite{freebase}, NELL \cite{nell} and Wikidata \cite{wikidata}. They are widely used as background knowledge provider in many applications such as natural language understanding \cite{DBLP:journals/corr/abs-1909-04164}, recommender system \cite{recsys}, and question-answering \cite{qa}.

\begin{table}[tbp]
    \small
    \centering
    \caption{Comparison of different KGE toolkits.}
    \vspace{-1mm}
    \label{tab:toolkit}
    \begin{tabular}{l|l|c|c|c}
    \hline
    \multicolumn{2}{c|}{}&C-KGE&GNN-based KGE&Rule-based KGE\\
    \hline
    \multirow{3}{*}{Input}&Triple&$\bullet$&$\bullet$&$\bullet$\\
    &Graph&&$\bullet$&\\
    &Rule&&&$\bullet$\\
    \hline
    \multicolumn{2}{l|}{OpenKE~\cite{han2018openke}} & \CheckmarkBold & &\\
    \multicolumn{2}{l|}{DGL-KE~\cite{DGL-KE}} & \CheckmarkBold& &\\
    \multicolumn{2}{l|}{Pykg2vec~\cite{yu2019pykg2vec}} & \CheckmarkBold &&\\
    \multicolumn{2}{l|}{AmpliGraph~\cite{ampligraph}} & \CheckmarkBold & &\\
    \multicolumn{2}{l|}{TorchKGE~\cite{arm2020torchkge}} & \CheckmarkBold && \\
    \multicolumn{2}{l|}{LIBKGE~\cite{libkge}} & \CheckmarkBold & &\\
    \multicolumn{2}{l|}{PyKEEN 1.0~\cite{ali2021pykeen}} & \CheckmarkBold&\CheckmarkBold &\\
    \multicolumn{2}{l|}{\textbf{NeuralKG}} & \CheckmarkBold & \CheckmarkBold & \CheckmarkBold \\
    \hline
    \end{tabular}
    \vspace{-5.8mm}
\end{table}

Traditional query and reasoning on KGs are based on manipulation of symbolic representations, which is vulnerable to the noisiness and incompleteness of KGs. Thus, with the development of deep learning, representation learning of KGs has been widely explored, aiming to embed a KG into a low-dimensional vector space while preserving the structural and semantic information contained in the KG. Thus these methods are also called Knowledge Graph Embeddings (KGEs). 
Flourishing research has been conducted in this area, and different series of KGEs are proposed. 
Conventional KGEs (C-KGEs)\cite{TransE,DistMult} apply a geometric assumption in vector space for true triples and use single triple as input for triple scoring. 
They ignore the rich graph structures of entities that reflect their semantics. Thus Graph Neural Network (GNN) effective in encoding graph structures are widely applied in KGEs. Based on the input graph, GNN-based methods\cite{RGCN,CompGCN} use representations of entities aggregated from their neighbors in the graph instead of embeddings of them for triple scoring, which help them capture the graph patterns explicitly. 
Apart from triples and graphs with triples, another essential part of KGs, logic rules, are also considered in KGEs. Rules define higher-level semantics relationships between different relations. Many Rule-based KGEs \cite{IterE,guo2017knowledge} tried to inject logic rules, either pre-defined or learned, into KGEs to improve the expressiveness of models. 
\begin{figure*}[tbp]
    \centering
\includegraphics[width=0.95\textwidth]{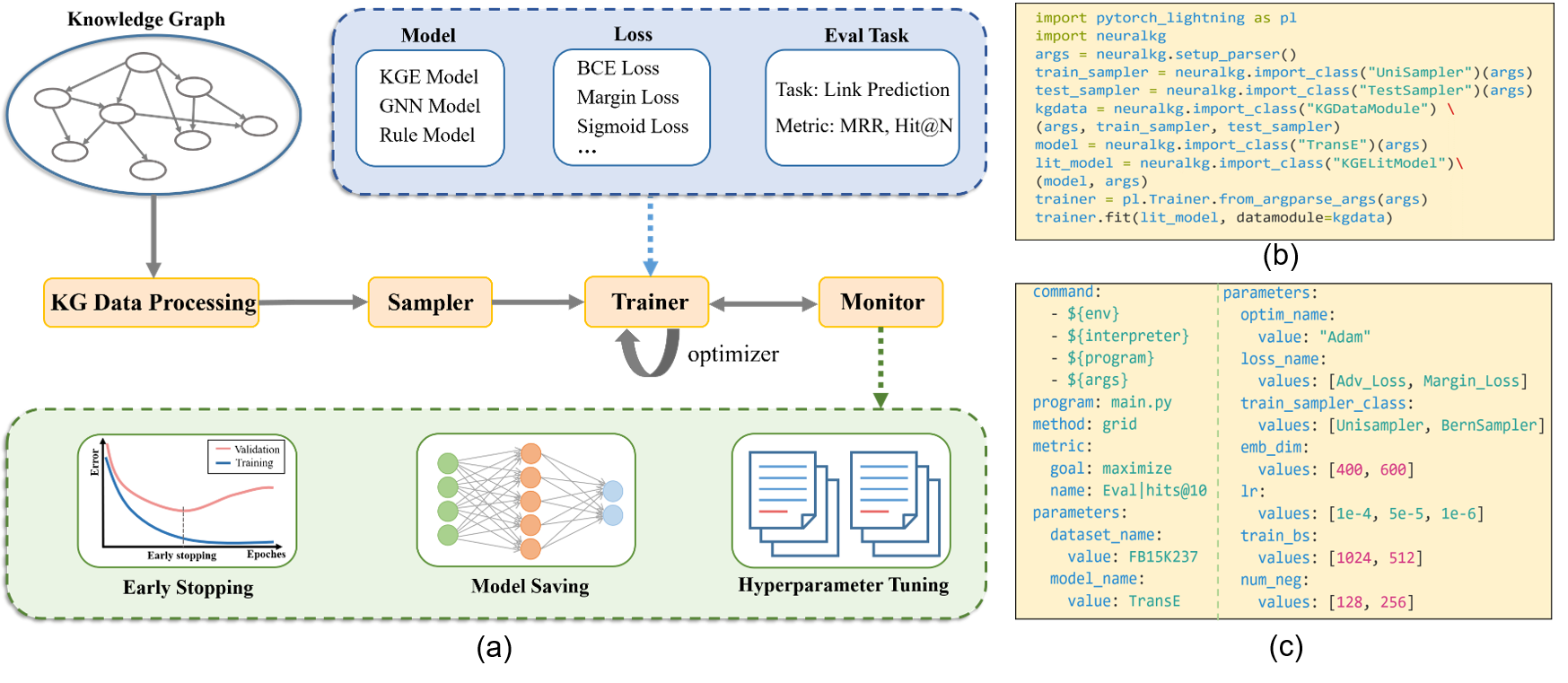}
	\caption{(a) The pipeline of NeuralKG. (b) Quick start example. (c)  Hyperparameter tuning example.}
	\vspace{-3mm}
	\label{fig:pipeline}
\end{figure*}

Although there are a lot of KGE methods and authors public their source code, it is difficult for others to apply and compare them due to the difference of programming languages (e.g., C++, Python, Java) and frameworks (e.g., TensorFlow, PyTorch, Theano). 
Thus several open-source knowledge graph representation toolkits have been developed, including OpenKE \cite{han2018openke}, Pykg2vec \cite{yu2019pykg2vec}, TorchKGE \cite{arm2020torchkge}, LIBKGE \cite{libkge}, and DGL-KE \cite{DGL-KE}. They provide implementations and evaluation results of widely-used and recently-proposed models, making it more easier for people to apply those methods.

However, most of these toolkits focus on the implementation C-KGEs, and a few of them provide the implementation for GNN-based KGEs, and none of them supports Rule-based KGEs. Thus for a more general purpose of diverse representation learning of KGs, it is necessary to build a toolkit that supports the implementation and creation of all three series of methods mentioned before. 
While building such a toolkit is non-trivial. These three series of methods are different but not independent. For example,  C-KGEs are the backbone of most GNN-based and Rule-based KGEs. Thus, decoupling components should be contained in the concise toolkit to train all methods in a unified framework. 

In this paper, we present such an open-source library for diverse representation learning of KGs. We named it \textbf{NeuralKG} targeting at general neural representations in vector spaces for KGs. As shown in Table \ref{tab:toolkit}, it supports the development and design of all three series of KGEs, C-KGEs, GNN-based KGEs, and Rule-based KGEs. 
It is a unified framework with various decoupled modules, including KG Data Preprocessing, Sampler for negative sampling, Monitor for hyperparameter tuning, Trainer covering the training, and model validation.
Thus users could utilize it for comprehensive and diverse research and application of representation learning on KGs. Furthermore, we provide detailed documentation for beginners to make it easy to use NeuralKG library. We also created a website of NeuralKG to organize an open and shared KG representation learning community. More importantly, we will also present long-term technical support to meet new needs in the future. 
In summary, the features of NeuralKG library are as follows: 
\begin{itemize}
    \item \textbf{Support diverse types of methods.} NeuralKG, as a library for diverse representation learning of KGs, provides implementations of three series of KGE methods, including C-KGEs, GNN-based KGEs, and Rule-based KGEs. 
    \item \textbf{Support easy customization.} NeuralKG contains fine-grained decoupled modules that are commonly used in different KGEs. With NeuralKG, users can develop their KGEs quickly and obtain optimal trained models and hyper-parameters. 
    \item \textbf{With detailed documentation and support.} The core team of NeuralKG provides detailed documentation, an online platform to organize a shared KG representation learning community, and long-term technical maintenance.
\end{itemize}





\section{NeuralKG library}
\subsection{Library Overview}
NeuralKG is built on PyTorch Lightning \footnote{\url{https://www.pytorchlightning.ai}}. It provides a general workflow of diverse representation learning on KGs and is highly modularized, supporting three series of KGEs.
Figure \ref{fig:pipeline} (a) shows the pipeline of NeuralKG, given a knowledge graph. Firstly, the KG data processing module processes the dataset into a format that facilitates data loading and constructs several dictionaries of data statistics. Secondly, the sampler module performs negative sampling on the dataset based on statistical information from the KG data processing module. Thirdly, the trainer module takes the positive and negative samples from the sampler module as input and conducts iterative training and validation of different models, where concrete representation learning models, loss functions, and evaluation tasks are integrated. At the same time, the monitor module observes the training status of models. It enables early stopping to prevent overfitting and avoid computation resources wastes, and supports hyperparameter tuning to help users obtain the optimally trained model.

\subsection{Library Modules}
\label{sec:library}
This section introduces the details of various scalable modules provided in NeuralKG. 
It includes KG Data Processing, Sampler, Trainer and Monitor. 

\subsubsection{KG Data Processing}

KG data processing module preprocesses the input KG dataset for data loading. It generates necessary statistics for subsequent negative sampling and supports the development of custom classes for personal data processing. It mainly includes \texttt{KGData}, \texttt{KGDataModule}, and \texttt{Grounding}.


\paragraph{KGData.} 
The \texttt{KGData} class loads raw KG data and assigns a unique numerical id, usually starting from $0$, for each entity and relation in the dataset. It also provides functions to generate data statistics required in negative sampling, including a function named \textit{get\_hr2t\_rt2h\_from\_train} which constructs a dictionary to store tail entities corresponding to a head entity and relation pair ($h$, $r$) in the training set, as well as head entities for each ($r$, $t$) pair, and \textit{count\_frequency} function counting the frequency of each pair.


\paragraph{KGDataModule.} 
The \texttt{KGDataModule} class is inherited from the \texttt{LightningDataModule} class in PyTorch Lightning, which combines the functionality of the \texttt{Dataset} and \texttt{Dataloader} classes in PyTorch. It supplies several interfaces for stacking the corresponding data into mini-batches for the training, validating , and testing phases in \texttt{Trainer} module.

\paragraph{Grounding.} The \texttt{Grounding} class reads rule dataset, provides the function \textit{groundRule} to generate groundings of each rule, that is replacing all variables in rules with concrete entities in the dataset, and saves all groundings to a file if necessary. It will be used in Rule-based KGEs.

\subsubsection{Sampler}
Sampler module includes 
\texttt{BaseSampler}, \texttt{RevSampler}, \texttt{NegSampler}, and \texttt{GraphSampler}.
All these samplers are general and independent of datasets and methods. Thus, users can flexibly use different samplers based on their targets.

\paragraph{BaseSampler/RevSampler.} 
\texttt{BaseSampler} and \texttt{RevSampler} are parent classes of \texttt{NegSampler} and \texttt{GraphSampler}.  
During negative sampling, they help filter generated negative samples that exist in the training dataset, according to the dictionary of statistics from \texttt{KGData} module. Furthermore, \texttt{RevSampler} also adds inverse relations into the training dataset, thus allowing models convert head entity prediction $(?, r,t)$ into tail entity prediction $(t, r^{-1}, ?)$, where $r^{-1}$ is the inverse relation of $r$.   
 

\paragraph{NegSampler.} 
The \texttt{NegSampler} class provides a variety of negative sampling methods given a triple $(h,r,t)$: 
\texttt{UniSampler} randomly replacing $h$ or $t$ following uniformly distribution~\cite{TransE}, 
\texttt{BernSampler} randomly replacing  $h$ or $t$ following a Bernoulli distribution created for each relation based on statistics from \texttt{KGData} module~\cite{TransH},
\texttt{AdvSampler} applying self-adversarial negative sampling proposed in \cite{RotatE}, and 
\texttt{AllSampler} replacing $h$ or $t$ with all entities in the datasets. 
In addition, \texttt{NegSampler} supports customization of negative samplers such as \texttt{CrossESampler} for CrossE\cite{CrossE}. 

\paragraph{GraphSampler.} 
 The \texttt{GraphSampler} class uses the graph of statistics built in DGL \cite{wang2019dgl} to sample triples and record the normalization of nodes and edges.  It uses uniform negative sampling to create negative samples. Similar to \texttt{Negsampler}, \texttt{GraphSampler} also supports the customization of graph samplers such as \texttt{KBATSampler} and \texttt{CompGCNSampler} for KBAT \cite{KBGAT} and CompGCN \cite{CompGCN} respectively.

\subsubsection{Trainer}
NeuralKG uses the trainer module to guide models' iterative training and validation. 
The detailed experimental guidance and model reproduction results are provided in NeuralKG, facilitating developers and researchers to quickly start and comprehensively compare different types of KGE models with NeuralKG.

\paragraph{Training.}
The trainer module specifies the training details of models and controls models' training progress. In the trainer module, loss function and optimizer could be specified and customized. During training, it helps display the training progress bar of models in real-time, and supports breakpoint retraining and saving training logs for subsequent experimental analysis. 

\paragraph{Evaluation.} 
For evaluation of models, such as getting predicting results on valid or test data, the trainer module accepts external parameters such as \textit{check\_per\_epoch} setting the number of rounds of model evaluation, and  \textit{limit\_val\_batches} for validating the model with a portion of the valid data to help users debug quickly during the implementation of new models. By default, Mean Reciprocal Rank (\textit{MRR}) and Hit at top K (\textit{Hit@K}) metrics are applied to evaluate the link prediction results of models. 


\subsubsection{Monitor}
Monitor module outputs the training status of the models, early stop models' training if there are continuous drops on user-specified evaluation metrics during evaluation. It saves the model with the best evaluation results. Moreover, it supports hyperparameter tuning to help users obtain the best hyperparameter setting regarding evaluation results.


\paragraph{Hyperparameter Tuning.}
The combination of different independent components and hyperparameters in the training pipeline has a significant impact on model performance~\cite{ruffinelli2020you}.
Therefore, NeuralKG utilizes Weights $\&$ Biases \footnote{\url{https://wandb.ai/site}} to provide grid search, random search, and Bayesian search for automatic hyperparameter optimization.
More importantly, NeuralKG treats everything as hyperparameters. 
Thus besides usual hyperparameters like learning rate, batch size, and embedding dimension, users also can search for standard components used in the representation learning model's pipeline such as negative sampling methods, loss functions, and optimizers. Users only need to modify the configuration file in the format "*.yaml" to perform hyperparameter optimization.
The example in Figure~\ref{fig:pipeline} (c) displays hyperparameter optimization of the TransE \cite{TransE} on the FB15K-237 \cite{fb15k237} dataset using grid search.

\subsection{Model Hub}
\label{sec:modelhub}
\begin{table}[tbp]
    \centering
    \caption{NeuralKG Model Hub.}
    \vspace{-1mm}
    \label{tab:model}
    \begin{tabular}{l|l}
    \hline
    \textbf{Components} &\multicolumn{1}{c}{ \textbf{Models}} \\
    \hline
    KGEModel & \multicolumn{1}{m{5.5cm}}{TransE~\cite{TransE}, TransH~\cite{TransH}, TransR~\cite{TransR}, ComplEx~\cite{ComplEx}, DistMult~\cite{DistMult}, RotatE~\cite{RotatE}, ConvE~\cite{ConvE}, BoxE\cite{ACLS-NeurIPS2020}, CrossE~\cite{CrossE}, SimplE~\cite{kazemi2018simple}}\\
    \hline
    GNNModel & \multicolumn{1}{m{5.5cm}}{RGCN~\cite{RGCN}, KBAT~\cite{KBGAT}, CompGCN~\cite{CompGCN}, XTransE~\cite{XTransE}} \\
    \hline
    \multirow{2}{*}{RuleModel} & ComplEx-NNE+AER~\cite{boyang2018:aer}, RUGE~\cite{guo2017knowledge}\\ &IterE~\cite{IterE} \\
    \hline
    \end{tabular}
    \vspace{-1mm}
\end{table}


This section introduces the model hubs of NeuralKG.  NeuralKG provides the implementation of three series of KGEs in three modules, KGEModel for C-KGEs, GNNModel for GNN-based KGEs, and RuleModel for Rule-based KGEs. All models in NeuralKG are listed in Table~\ref{tab:model}.
\subsubsection{KGEModel.}
This model hub covers a wide range of C-KGEs, including models that use distance-based functions to calculate sample scores, such as TransE \cite{TransE} and Rotate \cite{RotatE}, models that use a similarity-based approach to obtain sample scores such as Distmult\cite{DistMult}, and models use convolutional neural network such as ConvE \cite{ConvE}.
Notes that some models proposed in the last year, such as BoxE \cite{ACLS-NeurIPS2020}, are also included in NeuralKG, which are missing in other toolkits. In Table \ref{tab:transe}, we show the link prediction results of the most widely used and implemented model TransE on FB15k-237. 
Compared to existing toolkits, NeuralKG achieves reasonable results.
\begin{table}[tbp]
    \centering
    \caption{TransE Evaluation Results on FB15K237. $^{*}$ represents the reproduced result of using toolkits, $^{\Delta}$ indicates the reproduced result reported by toolkits.}
    \vspace{-1mm}
    \label{tab:transe}
    \begin{tabular}{ccccc}
    \hline
    Toolkits&MRR&HITS@1&HITS@3&HITS@10 \\
    \hline
    LIBKGE\cite{libkge}$^{\Delta}$ &0.31&0.22&0.35&0.50\\
    OpenKE\cite{han2018openke} &0.29$^{\Delta}$&0.19$^{*}$&0.32$^{*}$&0.48$^{\Delta}$\\
    DGL-KE\cite{DGL-KE}$^{*}$ &0.23&0.13&0.27&0.43 \\
    TorchKGE\cite{arm2020torchkge}$^{*}$&0.29&0.20&0.31&0.46\\
    Pykg2vec\cite{yu2019pykg2vec}$^{*}$&0.25&0.16&0.28&0.42\\
    AmpliGraph\cite{ampligraph}$^{\Delta}$ &0.31&0.22&0.35&0.50\\
    PyKEEN 1.0\cite{ali2021pykeen}$^{*}$&0.14&0.05&0.11&0.30 \\
    \hline
    NeuralKG&0.32&0.23&0.36&0.51\\
    \hline
    \end{tabular}
\end{table}

\subsubsection{GNNModel.} 
This model hub covers the common GNN-based KGEs, including RGCN \cite{RGCN}, KBAT  \cite{KBGAT}, CompGCN \cite{CompGCN} and XTransE \cite{XTransE}. GNNs have been proved to be efficient and effective in encoding graph structures, thus a series of works tried to implement GNN as encoders for entity representations and apply conventional KGEs as decoders for triple scoring. Table \ref{tab:rgcn} presents the link prediction results of the most widely used model RGCN \cite{RGCN} on FB15k-237, which shows that NeuralKG successfully achieves comparable results compared to results reported in the original paper.


\subsubsection{RuleModel.}
This model hub covers a series of methods that injects logic rules during representation learning of KGs, including ComplEx-NNE+AER \cite{boyang2018:aer}, RUGE \cite{guo2017knowledge}, IterE\cite{IterE}. 
Rules are used to regularize the relation embeddings or infer extra triples for training via grounding. 
With rule injected, KGEs earns a higher expressiveness regarding semantics. 
In Table~\ref{tab:ruge}, we show the link prediction results of RUGE~\cite{guo2017knowledge} from NeuralKG and from the original paper. They are comaprable. 


\subsubsection{Quick Start} 
As shown in Table~\ref{tab:model}, \ref{tab:transe}, \ref{tab:rgcn} and \ref{tab:ruge}, NeuralKG provides implementations and evaluation results of widely used models in three genres. 
With a unified model library, NeuralKG supports users to quickly build their models. An example is shown in Figure~\ref{fig:pipeline} (b).

\begin{table}[tbp]
    \centering
    \caption{RGCN Evaluation Results on FB15K237.}
    \vspace{-1mm}
    \label{tab:rgcn}
    \begin{tabular}{ccccc}
    \hline
    &MRR&HITS@1&HITS@3&HITS@10 \\
    \hline
    RGCN-paper\cite{RGCN} &0.25&0.15&0.26&0.41\\
    NeuralKG &0.25&0.16&0.27&0.43\\
    \hline
    \end{tabular}
\end{table}
\begin{table}[tbp]
    \centering
    \caption{RUGE Evaluation Results on FB15K.}
    \vspace{-1mm}
    \label{tab:ruge}
    \begin{tabular}{ccccc}
    \hline
    &MRR&HITS@1&HITS@3&HITS@10 \\
    \hline
    RUGE-paper\cite{guo2017knowledge} &0.77&0.70&0.82&0.87\\
    NeuralKG &0.76&0.70&0.81&0.85\\
    \hline
    \end{tabular}
    \vspace{-1mm}
\end{table}

\section{Related Work}
With the widespread use of knowledge graphs in recent years, 
many open-source KGE toolkits for reproducible research have emerged in the knowledge representation learning community. 
OpenKE is an early open-source framework for KGEs, backed by both TensorFlow and PyTorch libraries and some features implemented in C++.
DGL-KE\cite{DGL-KE} is a scalable package for learning large-scale knowledge graph embeddings and supports multi-GPU training. AmpliGraph\cite{ampligraph} and Pykg2vec\cite{yu2019pykg2vec} are released at the same time in 2019, which focus on providing intuitive APIs to facilitate effective training of KGE models. 
PyKEEN\cite{ali2021pykeen} 1.0 is a re-designed and re-implemented toolkit for the PyKEEN toolkit. Thanks to the community's efforts, it provides automatic memory optimization modules to utilize the hardware resource as much as possible.  Other toolkits, LIBKGE\cite{libkge} and TorchKGE\cite{arm2020torchkge}, further help to build a modular framework to explore the impact of different components of the KGE pipeline and get quick evaluation results. 
However, these toolkits focus on the implementation of conventional KGEs which ignore other types of methods, thus are not a general library for representation learning of knowledge graphs. 
\section{Conclusion}
With the development of neural knowledge graph representation learning, different kinds of KGE models have been proposed, whose performance in practical applications is affected by negative sampling mechanisms, model design, and training strategies. In this work, we introduce NeuralKG, an open-source and extensible knowledge graph representation learning library. It uses a unified framework to implement three types of  KGE methods, including conventional KGEs, GNN-based KGEs, and Rule-based KGEs, showing a more comprehensive view of KGE methods than existing toolkits. 
To meet users' customization needs, it contains decoupled modules, helping users to develop their models and get the best training methods for research and applications on knowledge graph representation learning. 
In addition, we offer an online platform around this project to organize an open and shared community.
In the future, the core team of NeuralKG will continuously provide implementations of new KGE methods of different types. At the same time, we'll try to enable the crowdsourcing of models from other developers. 



\bibliographystyle{ACM-Reference-Format}
\bibliography{citation}


\appendix

\end{document}